# Image inpainting enhancement by replacing the original mask with a self-attended region from the input image

Kourosh Kiani[1*], Razieh Rastgoo[1], Alireza Chaji[1], Sergio Escalera[2]

Electrical and Computer Engineering Department, Semnan University, Semnan, 3513119111, Iran

Department of Mathematics and Informatics, Universität de Barcelona, and Computer Vision Center, Barcelona, Spain

[*]Corresponding Author (Kourosh.kiani@semnan.ac.ir)

**Abstract.** Image inpainting, the process of restoring missing or corrupted regions of an image by reconstructing pixel information, has recently seen considerable advancements through deep learning-based approaches. In this paper, we introduce a novel deep learning-based pre-processing methodology for image inpainting utilizing the Vision Transformer (ViT). Our approach involves replacing masked pixel values with those generated by the ViT, leveraging diverse visual patches within the attention matrix to capture discriminative spatial features. To the best of our knowledge, this is the first instance of such a pre-processing model being proposed for image inpainting tasks. Furthermore, we show that our methodology can be effectively applied using the pre-trained ViT model with pre-defined patch size. To evaluate the generalization capability of the proposed methodology, we provide experimental results comparing our approach with four standard models across four public datasets, demonstrating the efficacy of our pre-processing technique in enhancing inpainting performance.

**Keywords:** Image inpainting, Generative Adversarial Network (GAN), Vision Transformer (ViT), Loss, Reconstructed image.

## 1. Introduction

Image inpainting, commonly referred to as image completion or reconstruction, involves estimating pixel values to restore missing regions within an input image. As an interesting task in computer vision, image inpainting supports numerous applications, including image editing (Xiaobo Zhang, 2023), image-based rendering (Weize Quan, 2024), computational photography (Xiaobo Zhang, 2023), object removal (Weize Quan, 2024), and image denoising (Jiahui Yu, 2018). One of the primary challenges in this field is synthesizing visually realistic and semantically plausible pixels for missing regions that harmonize with the surrounding content (Jiahui Yu, 2018). To address this challenge, researchers have proposed various solutions in recent years (Yi Wang, 2018; Alexei A. Efros W. T., 2001; Alexei A. Efros T. K., 1999; Cho, Butman, Avidan, & Freeman, 2008; Kaiming He, 2016; Ishaan Gulrajani, 2017; Omar Elharrouss; Satoshi Iizuka, 2017; Yijun Li, 2017; Deepak Pathak, 2016) (Raymond A. Yeh, 2017). However, there remains significant potential to enhance the quality of generated images. One promising approach involves leveraging auxiliary information, either from surrounding areas within the image or external data sources (Yi Wang, 2018). Another strategy adapts techniques from texture synthesis, in which background patches are matched and transferred into missing regions, either progressing from low-resolution to high-resolution or propagating from hole boundaries (Alexei A. Efros W. T., 2001; Alexei A. Efros T. K., 1999). This technique is particularly effective in background inpainting tasks and is widely applied in practical contexts (Cho, Butman, Avidan, & Freeman, 2008). However, it struggles with complex, non-repetitive structures and faces challenges in capturing high-level semantic information.

Recent advancements in deep learning, particularly through Convolutional Neural Networks (CNNs) (Kaiming He, 2016), Generative Adversarial Networks (GANs) (Ishaan Gulrajani, 2017), and Transformer models (Omar Elharrouss), have led to substantial progress in image inpainting (Satoshi Iizuka, 2017; Yijun Li, 2017; Deepak Pathak, 2016; Raymond A. Yeh, 2017). In this framework, image inpainting is formulated as a conditional image generation problem, utilizing a convolutional encoder-decoder network trained alongside adversarial networks to ensure consistency between generated and real pixels. While these models generate plausible new content, they often exhibit boundary artifacts, distorted structures, and blurry textures that are inconsistent with surrounding areas. Consequently, despite the advancements in inpainting models, further improvements are needed to achieve enhanced performance in this field.

In general, an image inpainting task aims to address three problems: feature extraction, finding neighbor patches, and collecting auxiliary information. The first problem aims to extract the effective features for making connections between missing and known areas. Relying on automatic feature extraction from data, deep learning-based models have been extensively used in recent years for this end. One of the most interesting deep learning-based models for image inpainting is the encoder-decoder model that extracts the features using the CNN in both encoder and decoder parts of the model (Jiahui Yu, 2018; Satoshi Iizuka, 2017; Deepak Pathak, 2016; Raymond A. Yeh, 2017; Chao Yang, 2017). More specifically, Pathak et al. presented Context Encoders, as a CNN-based model trained to generate the contents of an arbitrary image region conditioned on its surroundings, for unsupervised visual feature learning in the image inpainting task. Considering both the content of the entire image as well as plausible contents for the missing parts of the image, the proposed model can successfully reconstruct the image. More specifically, the Context Encoders model can simultaneously learn the appearance and also the semantics of visual structures in the image (Deepak Pathak, 2016). Moreover, a multi-scale neural patch synthesis method is proposed by Yang et al. (Chao Yang, 2017) based on joint optimization of image content and texture constraints. This method aims to keep both contextual structures as well as high-frequency details of the image. Results on two public datasets show that this model can produce sharper and more coherent results than prior methods (Chao Yang, 2017). Following the (Deepak Pathak, 2016; Chao Yang, 2017), a new deep generative model has been designed in (Jiahui Yu, 2018) to synthesize novel image structures using surrounding image features. The proposed model contains a fully convolutional neural network for processing multiple holes with variable sizes at different locations in the image. Results on multiple datasets confirm the effectiveness of the proposed methodology. Thought, these models need to simultaneously consider and combine both global and local features in the model to enhance the results. The second problem is conducted to explicitly finding the neighbor components in the image for generating the realistic details (Jiahui Yu, 2018; Satoshi Iizuka, 2017; Deepak Pathak, 2016; Raymond A. Yeh, 2017; Chao Yang, 2017) (Kaiming He J. S., 2012; Connelly Barnes, 2009). Complex and various structures in the missing areas and the context can lead to the performance degradation of the generation process. Moreover, the process of finding the neighbor patches is time-consuming. To overcome this challenge, the proposed model in this work applies the search mechanism only during the train phase for finding the neighbor patches. Merging the auxiliary information to make the optimal candidates for missing patches is the main idea of the third problem. Using the spatial-variant constraints can help to make the optimal patch candidates by assigning the lower and higher constraints to the boundary and center areas, respectively. In this way, the adversarial loss has been recently employed to learn multi-modality by assigning different weights to loss for boundary consistency (Connelly Barnes, 2009; Kaiming He J. S., 2014). In addition, the multi-column structure (Dan Ciregan, 2012; Yingying Zhang, 2016; Forest Agostinelli, 2013) is used in the model since it can decompose images into components with different receptive fields and feature resolutions. Unlike multi-scale or coarse-to-fine strategies (Chao Yang, 2017; T. Karras, 2018) that use resized images, branches in the multi-column network directly use full-resolution input to characterize multi-scale feature representations regarding global and local information. Moreover, an Implicit Diversified Markov Random Field (ID-MRF) term is used in the training phase only. Rather than directly using the matched features, which may lead to visual artifacts, this term is incorporated as a regularization term. Additionally, a confidence-driven reconstruction loss is employed that constrains the generated content according to the spatial location. With all these improvements, the performance improvement is obtained using these methods.

In image inpainting, preprocessing plays a crucial role in improving the quality of the output. Preprocessing helps to prepare the input images, ensuring that the inpainting model performs optimally. There are different preprocessing mechanisms that can be used in an image inpainting task, such as normalization, mask creation, denoising, and edge detection (Xiaobo Zhang, 2023), enhancing the model's ability to produce high-quality inpainting results. These preprocessing techniques are often adjusted based on the specific dataset and the model architecture used in the inpainting task. In this way, we focus on the inpainting mask, aiming to enrich it before feeding to an image inpainting model. To this end, we propose a pre-processing methodology using Vision Transformer (ViT) model and various visual patches in the image. More specifically, our contributions can be summarized as follows:

- **Pre-processing Mechanism**: ViT is used as a preprocessor for the input image. The intuition behind using ViT is substituting the mask values with the values obtained from the ViT. To this end, different visual patches are used in the input image, aiming to obtain discriminative spatial features. To the best of our knowledge, this is the first time that such a pre-processing model is proposed to the image inpainting task.

- **Performance**: Experimental results comparing with four standard models on four public datasets confirm the efficacy of the proposed pre-processing methodology for image inpainting task.

The rest of this paper is organized as follows: section 2 briefly reviews recent works in image inpainting. Details of the proposed model will be presented in section 3. Experimental results with presenting a brief introduction to datasets, implementation details, and a discussion on the obtained results are mentioned in section 4. Finally, section 5 concludes the work by providing a future roadmap for improvement.

## 2. Related work

Generally, the current models for image inpainting can be studied from different perspectives. For instance, some models employ the traditional methods as well as the low-level features to transfer the information from the background regions to the missing ones. However, these methods are more suitable for the stationary textures compared to non-stationary data such as natural images (C. Ballester, 2001; Marcelo Bertalmio, 2000 ). In this way, a bidirectional patch similarity-based method has been suggested by Simakov et al. (Denis Simakov, 2008) for modeling the nonstationary visual data in image inpainting. This model suffers from the computational complexity of patch similarity. To overcome this challenge, PatchMatch, as a fast nearest-neighbor method, has been suggested, obtaining the significant results in image inpainting (Connelly Barnes, 2009). The recent advances in deep learning models, especially CNN-based models, are used in some models for pixel prediction of the missing regions. In this way, some efforts have been done to developing the GAN-based models with the embedded CNN in the generator and discriminator networks. Different approaches have been used in this way, such as training the small regions using CNN (Rolf Köhler, 2014; Li Xu, 2014) and using the Context Encoders (Deepak Pathak, 2016) for inpainting large missing regions. Moreover, using the global and local discriminators as adversarial losses is the main idea of the model proposed by Iizuka et al. (Satoshi Iizuka, 2017) to improve the performance of the Context Encoders models. To this end, the dilated convolutions are employed to substitute channel-wise fully connected layer in Context Encoders, aiming to extend the receptive fields of output neurons. In addition, some studies have been concentrated on generative face inpainting. For instance, Yeh et al. (Raymond A. Yeh, 2017) suggested a model to find the nearest encoding in latent space of the image with missing regions and decode to obtain the completed image. Moreover, an auxiliary loss has been included in the loss function by Li et al. [12] for face completion. However, these models need post processing steps, such as image blending operation to enforce color consistency near the boundaries of the missing regions. Another approach defines the image inpainting task as an optimization problem using the ideas from the image stylization (Snelgrove, 2017; Chuan Li, 2016). In this way, a multiscale neural patch synthesis model has been designed by Yang et al. (Chao Yang, 2017) using the joint optimization of image content and texture constraints. While this model has obtained the promising results, it suffers from the high complexity due to the optimization process. Using the spatial attention in deep networks is another approach for learning the contextual information, aiming to improve the image inpainting performance. In this way, a parametric spatial attention approach, namely Spatial Transformer Network (STN), as well as the spatially attentive or active convolutional kernels (Yunho Jeon, 2017; Jifeng Dai, 2017) have been suggested by researchers for performance improvement in image inpainting task. However, these methods are not effective for modeling patch-wise attention as well as predicting a flow field from the background region to the hole.

Recent studies highlight the importance of capturing long-range dependencies in image inpainting tasks. To address this, many existing methods leverage attention mechanisms or transformers, typically at low resolutions to manage computational costs. In this way, Li et al. (Wenbo Li, 2022), have proposed a transformer-based model designed for large-hole inpainting, which integrates the strengths of reconstructed images. This model introduces a specialized inpainting-focused transformer block, where the attention mechanism selectively aggregates non-local information from a subset of valid tokens, guided by a dynamic mask. Experimental results show that the effectiveness of the proposed approach across multiple benchmark datasets. However, DL-based inpainting methods often suffer from artifacts, particularly around boundaries and in highly textured regions. To address these issues, Wu et al. (Haiwei Wu, 2021) have developed an end-to-end, two-stage generative model that operates in a coarse-to-fine manner. This approach combines a local binary pattern (LBP) learning network with an image inpainting network. In the first stage, a U-Net-based LBP learning network is employed to accurately predict the structural details of the missing regions, which then guides the second-stage inpainting network for more precise pixel restoration. Additionally, an enhanced spatial attention mechanism has been integrated into the inpainting network, ensuring consistency not only between the known and generated regions but also within the generated area itself. Evaluation results on public datasets demonstrate the effectiveness of the proposed model in (Haiwei Wu, 2021).

Aiming to make the performance improvement in image inpainting task, in this paper, we propose a pre-processing methodology using deep learning models. More specifically, we use the ViT model with various visual patches, aiming to fill the zero values in the missing areas with the values obtained from the ViT. We assess the generalization capability of the proposed methodology on four comparative models using four public datasets, confirming the efficacy of the proposed pre-processing methodology for image inpainting task.

## 3. Proposed approach

In this section, we present the details of the proposed approach for image inpainting, including two main blocks: ViT pre-processing and replacing the missing regions.

### 3.1. ViT pre-processing

Let consider an input image along with a binary region mask, M, which have to feed to an image inpainting model. Generally, most of the previous works use a binary mask, which includes a matrix filled with 0 and 1 values for the known and unknown pixels. Furthermore, the unknown regions are filled with zero values in input image. These models aim to complete the unknown regions of the input image and provide a complete image. Here, we propose a pre-processing methodology to replace the binary mask with an attended mask obtained from the ViT model. To this end, we use the input image, including the unknown regions filled with zero values and feed it to the ViT model. Relying on the self-attention mechanism in ViT, a feature map is obtained from the input image. Considering the task and the characteristics of the image data, different visual patches in the image can be used to obtain the features. Here, we consider three kinds of visual patches in the input image (vertical, horizontal, and square) to construct the self-attention matrix. More concretely, details of these patches are as follows:

- **Vertical patches**: In this approach, we use the vertical patches in the image to feed to the self-attention mechanism. Fig. 1 shows a sample image including the vertical patches.

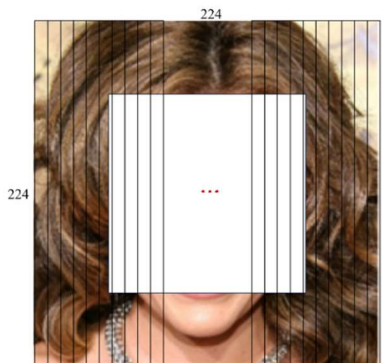

Fig. 1: Vertical patches in the input image.

The intuition behind using the vertical patches is capturing vertical features, such as buildings, trees, and other objects that stretch upward in images. Moreover, this kind of patches is efficient for obtaining the contextual information from the objects or patterns that are aligned vertically, making it suitable for tasks where vertical alignment is significant, like facial recognition (capturing nose, mouth, eyes, etc.). Considering these advantages, the vertical patches in Fig. 1 are self-attended to obtain the visual features from the image.

- **Horizontal patches**: In this approach, we use the horizontal blocks in the image to feed to the self-attention mechanism. Fig. 2 shows a sample image including the horizontal patches.

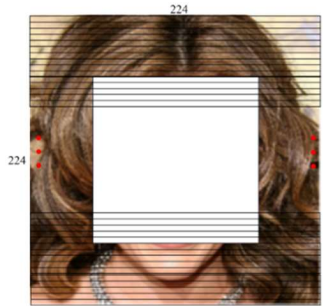
Fig. 2: Horizontal patches in the input image.

Horizontal patches capture horizontal features like landscapes, horizons, or wide objects. This is advantageous for tasks where the spatial relations across the width are important. Moreover, in tasks involving panoramic views or wide scenes (e.g., road images, landscapes), horizontal patches allow the model to capture wide features more efficiently. Considering these advantages, the horizontal patches in Fig. 2 are self-attended to obtain the visual features from the image.

- **Square patches**: In this approach, we use the square patches in the image to feed to the self-attention mechanism. Fig. 3 shows a sample image including the square patches.

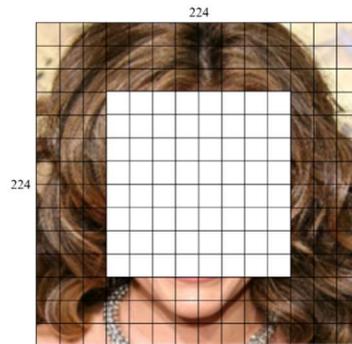
Fig. 3: Square patches in the input image.

Using square patches for self-attention in image processing, as commonly used in ViT model, has several advantages, such as balanced feature representation due to dividing the image evenly, allowing for balanced extraction of horizontal and vertical features. This symmetry ensures that both directions (horizontal and vertical) are treated equally, making it a robust choice for a wide variety of image types, whether they contain tall or wide objects. Additionally, square patches offer a uniform distribution across the image, making it ideal for tasks where no particular direction dominates, such as natural scenes, medical images, or textures. Considering these advantages, we use the square patches with different dimensions in the image to input to the self-attention mechanism.

### 3.2. Mask replacing

After pre-processing the input image using any of the vertical, horizontal, or square patches, the obtained feature map is used to multiply with the binary mask to fill the missing regions with the self-attended features from the ViT. In other words, instead of filling the missing regions with the zero values, we replace these regions with the attended features from the ViT. In this way, any image inpainting model can use this image as a more informative input, expecting to better refining the image. Fig. 4 shows the process of mask replacing in the proposed pre-processing methodology. Since the best results have been obtained using the vertical (column) attention matrix, we only visualize the results corresponding to this attention matrix.

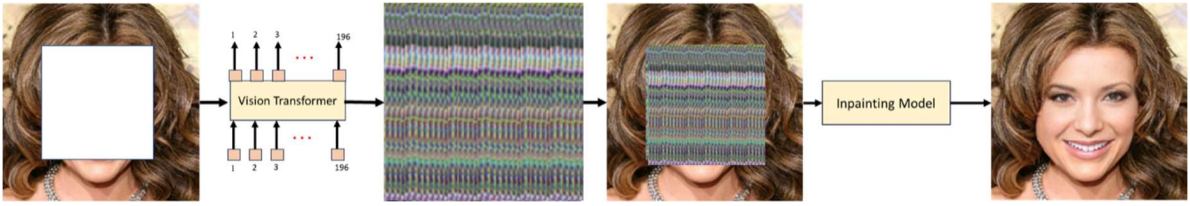
Fig. 4: The process of mask replacing in the proposed pre-processing methodology.

### 3.3. Training process

Training process includes the following steps:

- **Input Image** (Y): The input image, referred to as (Y), is considered to feed to the proposed model.
- **Binary Mask** (M): Generally, in an inpainting model, a binary mask is created in such a way that the value 0 indicates known pixels and the value 1 indicates unknown pixels. This mask is sampled at a random location on the image.
- **Masked Input Image** (X): Using the binary mask (M), a new image (X) is produced as follows:
$$X = Y \cdot (1 - M) \quad (1)$$
This operation keeps the known pixels from Y (because multiplying by 1 doesn't change the value) and sets the unknown pixels to 0 (since (1-M) will be 0 for unknown pixels).
- **Rich features extraction** (R): Relying on the ViT capabilities, richer visual features are extracted from the X, as follows:
$$X_{ViT} = ViT(X) \quad (2)$$
- **Model Input**: The generator model G takes the concatenation of $X_{ViT}$ and M as input. This means the model gets both the partially known image and the mask indicating where the unknown regions are.
$$X_R = X_{ViT} \cdot M + Y \cdot (1 - M) \quad (3)$$
- **Final Prediction** ($\hat{Y}$): The model generates a prediction for the unknown pixels. The final reconstructed image ($\hat{Y}$) is given by:
$$\hat{Y} = X_{ViT} \cdot M + Y \cdot (1 - M) + G(X_R, M) \odot M \quad (4)$$
The first and the second terms of above equation retain the known parts of the original image and the unknown parts with the model's predictions, respectively. G is a general inpainting model.

## 4. Experiments with four standard models

In this section, we present the experimental results of the proposed methodology on four comparative models on four datasets. We used four models with publicly available implementation. It is worth mention that the results have been obtained using the best pre-processing methodology (ViT with a 2-column attention matrix). Our ablation analysis on the proposed methodology will be presented in the next section.

### 4.1. GMCNN

The GMCNN (Yi Wang, 2018) is a generative multi-stream network for image inpainting, which synthesizes different components of an image in parallel within a single stage. To better capture global structures, a confidence-driven reconstruction loss as well as an implicit diversified Markov Random Field (MRF) regularization have been used to enhance local details. The combination of the multi-column network with the reconstruction and MRF losses allows for the effective propagation of both local and global context to the inpainting regions. An overview of this model has been shown in Fig. 5.

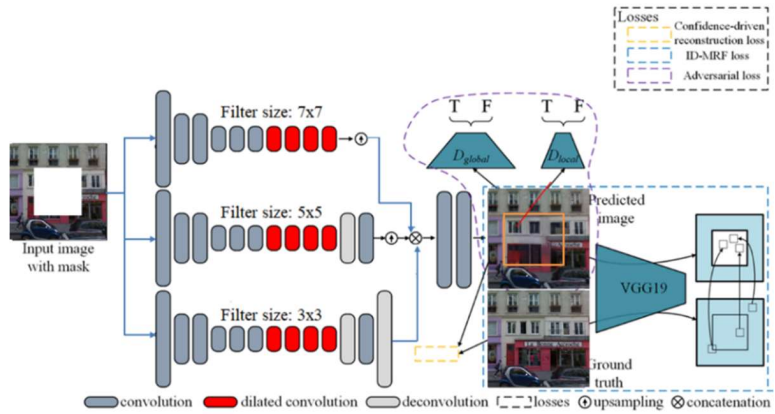

Fig. 5: An overview of the GMCNN model, including three convolution streams in an adversarial framework.

Considering our pre-processing methodology, the ViT model is used before feeding the input image to the model (Fig. 6).

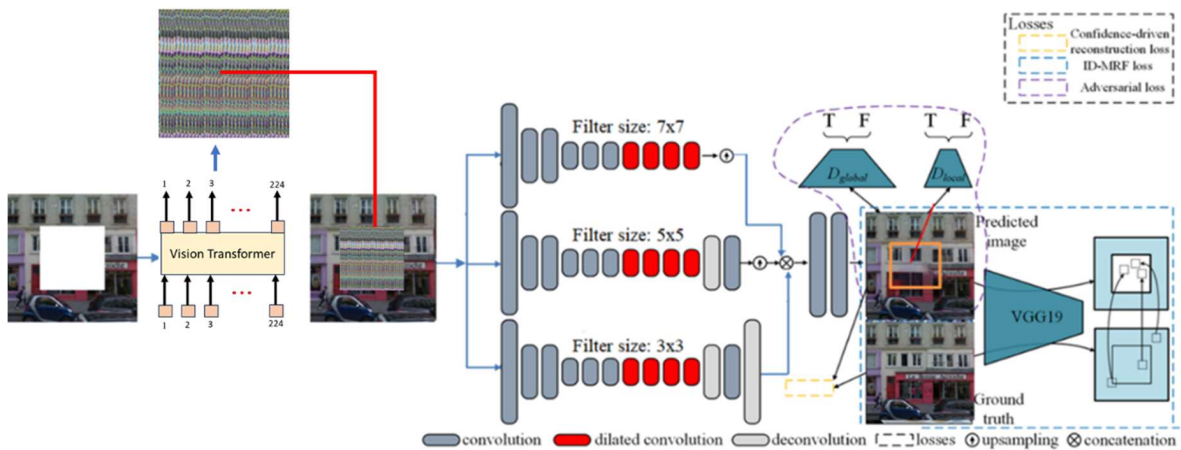

Fig. 6: Using the proposed pre-processing mechanism before feeding the input image to the model in [15].

## 4.2. MSNPS

MSNPS (Chao Yang, 2017) is a multi-scale neural patch synthesis approach that jointly optimizes image content and texture constraints (Fig. 7). This method not only maintains contextual structures but also generates high-frequency details by aligning patches with the most similar mid-layer feature correlations from a deep classification network. Testing on the ImageNet and Paris Streetview datasets, this model achieves state-of-the-art performance in inpainting accuracy.

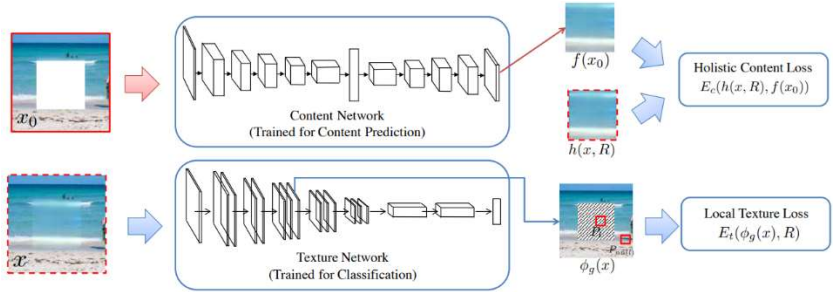

Fig. 7: An overview of the MSNPS model for image inpainting.

Using the proposed pre-processing methodology, the MSNPS model will be as follows (Fig. 8):

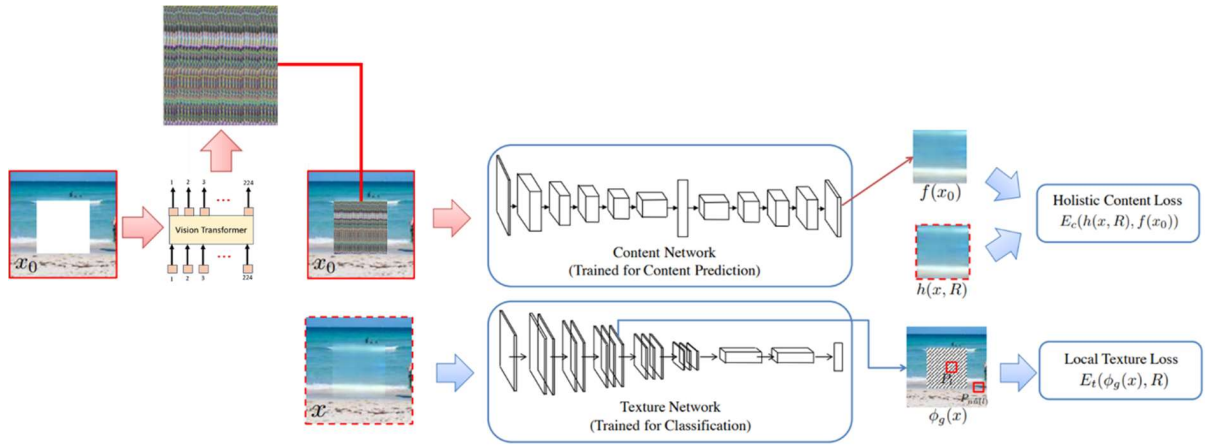

Fig. 8: Using the proposed pre-processing mechanism before feeding the input image to the model in [4].

4.3. CA

CA (Jiahui Yu, 2018) is a deep generative model that not only synthesizes new image structures but also explicitly leverages surrounding image features as references during training for more accurate predictions. This model is a fully convolutional, feed-forward neural network capable of handling multiple holes of varying sizes and arbitrary locations during testing. Experiments conducted on diverse datasets, including faces (CelebA, CelebA-HQ), textures (DTD), and natural images (ImageNet, Places2), demonstrate that this approach produces higher-quality inpainting results compared to existing methods. An overview of the CA model before and after applying the proposed pre-processing method can be found in Figs. 9 and 10, respectively.

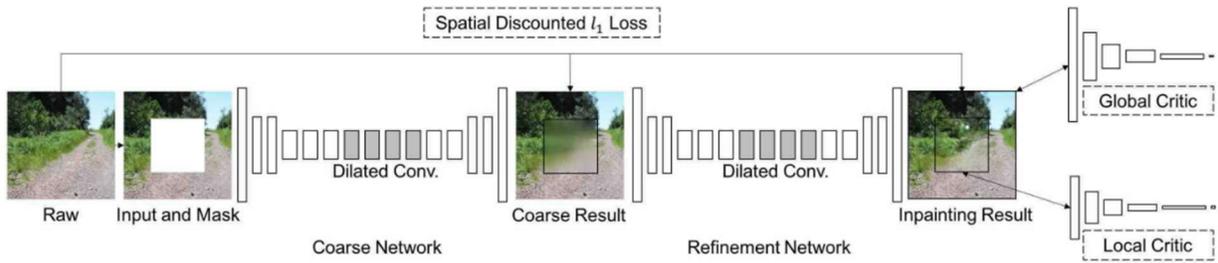

Fig. 9: An overview of the CA model *(Jiahui Yu, 2018)* for image inpainting.

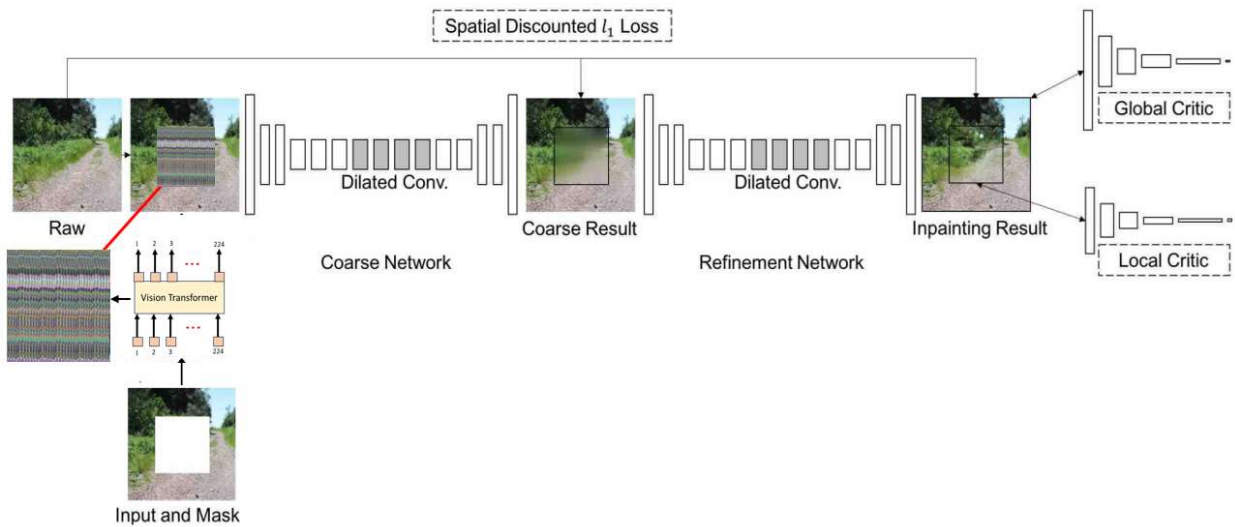

Fig. 10: Using the proposed pre-processing mechanism before feeding the input image to the model in (Jiahui Yu, 2018).

## 4.4. Context Encoders (CE)

CE is unsupervised visual feature learning algorithm based on context-driven pixel prediction. Inspired by the concept of autoencoders, CE is designed by the Authors in (Deepak Pathak, 2016), as a CNN designed to generate the contents of missing regions in an image, using the surrounding areas as context. To perform this task effectively, the network must not only comprehend the entire image's content but also generate plausible hypotheses for the missing parts. In training phase, two approaches are used: a standard pixel-wise reconstruction loss and a combination of reconstruction loss and adversarial loss. The latter yields significantly sharper results by better addressing the multimodal nature of the output. The experiments demonstrate that CE learn a representation that captures both the visual appearance and the underlying semantics of image structures. An overview of the CE model (Deepak Pathak, 2016) before and after applying the proposed pre-processing method can be found in Figs. 11 and 12, respectively.

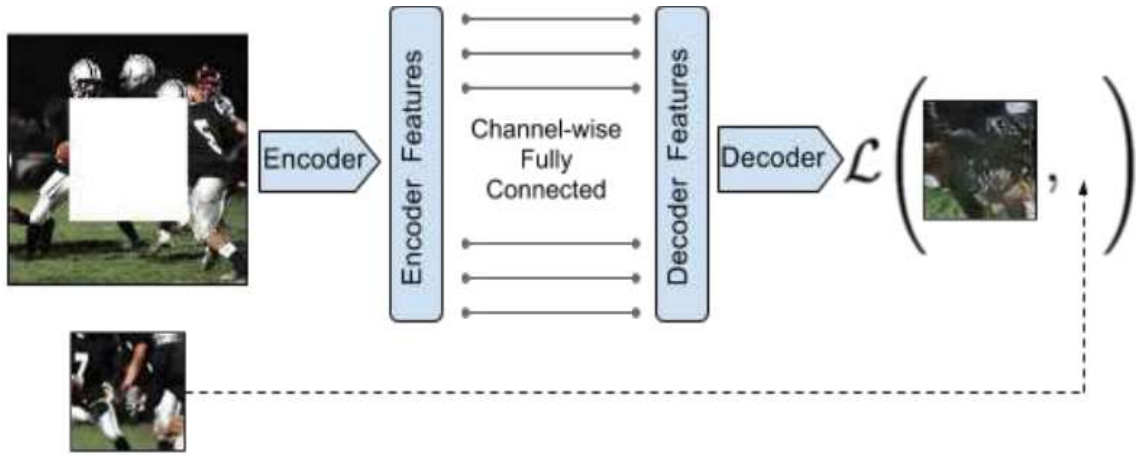

Fig. 11: An overview of the CE model in (Deepak Pathak, 2016) for image inpainting.

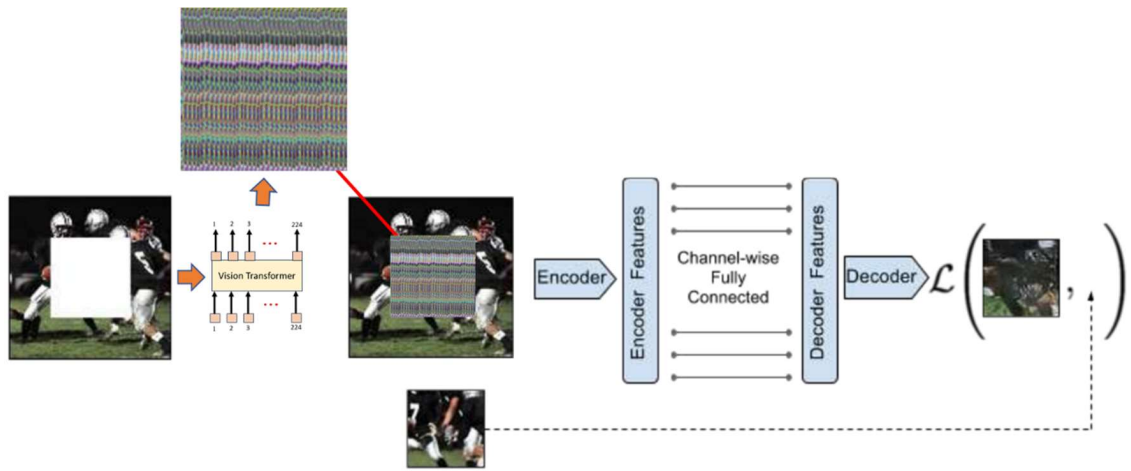

Fig. 12: Using the proposed pre-processing mechanism before feeding the input image to the model in (Deepak Pathak, 2016).

## 5. Experimental Results and Discussion

In this section, we delve into the results of the proposed model on four datasets. After the presentation of the implementation details of the proposed model, an overview of the datasets as well as the evaluation metrics, along with the ablation analysis, are presented. The section is concluded with a comparison with four comparative models, accompanied by a discussion on the obtained results.

### 5.1. Implementation Details

The implementation of our model utilizes the Python programming language and PyTorch library (PyTorch, 2024). PyTorch is a library specifically designed for data science and deep learning computations. The proposed model has been trained on a NVIDIA Tesla K80 GPU, employing the Adam optimizer, a mini-batch size of 64, a learning rate of 1e-4 with adaptive tuning, 300 epochs with early stopping, and a weight decay of 1e-5. For model evaluation, a subset of four datasets, namely Paris Street View (Deepak Pathak, 2016), Places2 (Places2, 2024), ImageNet (ImgNet, 2024), and CelebA-HQ (T. Karras, 2018), have been employed with the largest hole size 128 × 128 in random positions of the input images.

## 5.2. Datasets

Four datasets have been used for the model evaluation. Here, a brief introduction of these datasets is presented as follows:

- **Paris Street View**: This dataset includes approximately 10,000 images of 12 cities from two perspectives and the shape of 936 × 537 pixels.
- **Places2**: This dataset contains 10 million scene photographs, labeled with scene semantic categories, including a large and diverse list of the types of environments encountered in the world.
- **ImageNet**: There are 3.2 million images in total in 1000 categories in this dataset.
- **CelebA-HQ**: This dataset includes ten thousand identities, each of which has twenty images (two hundred thousand images in total).

## 5.3. Evaluation metrics

Two evaluation metrics are used to evaluate the model performance. A brief introduction to these metrics is as follows:

- **Peak Signal-to-Noise Ratio (PSNR)** (Alain Horé, 2010)**:** PSNR is a measure employed to quantify the quality of a reconstructed or compressed image compared to its original version. It is expressed in decibels (dB) and is derived from the Mean Squared Error (MSE) between the two images.
- **Structural Similarity Index Measure (SSIM)** (Alain Horé, 2010)**:** SSIM is a metric used to assess the visual impact of changes in structural information, luminance, and contrast between the original and a distorted image. Unlike PSNR, which focuses on pixel-wise errors, SSIM considers changes in structural information, making it more aligned with human visual perception.

## 5.4. Results and discussion

In this sub-section, we present the numerical and visual results obtained from the proposed pre-processing methodology. In this way, we experimented different ablation analysis on the GMCNN model, presenting 10 models with different components. Table 1 shows the descriptions of the experimented models in this way. In addition, results of these models have been shown in Table 2.

Table 1: Descriptions of the models used during the ablation analysis on the GMCNN model.

| Method | Description |
|---|---|
| GAN-VGG-16x16 | A model including the GAN, VGG, and the 16x16 attention matrix. |
| GAN-ResNet-16x16 | A model including the GAN, ResNet, and the 16x16 attention matrix. |
| GAN-VGG-ViT-16x16 | A model including the GAN, VGG, ViT, and the 16x16 attention matrix. |
| GAN-ResNet-ViT-16x16 | A model including the GAN, ResNet, ViT, and the 16x16 attention matrix. |
| GAN-ResNet-ViT(1-Column) | A model including the GAN, ResNet, ViT, and a 1-column attention matrix. |
| GAN-ResNet-ViT(2-Column) | A model including the GAN, ResNet, ViT, and a 2-column attention matrix. |
| GAN-ResNet-ViT(4-Column) | A model including the GAN, ResNet, ViT, and a 4-column attention matrix. |
| GAN-ResNet-ViT(1-Row) | A model including the GAN, ResNet, ViT, and a 1-Row attention matrix. |
| GAN-ResNet-ViT(2-Row) | A model including the GAN, ResNet, ViT, and a 2-Row attention matrix. |
| GAN-ResNet-ViT(4-Row) | A model including the GAN, ResNet, ViT, and a 4-Row attention matrix. |

Table 2: Numerical results of the ablation analysis on the GMCNN on four datasets.

| Method | Pairs street view-100 | | ImageNet-200 | | Places2-2K | | CelebA-HQ-2K | |
|---|---|---|---|---|---|---|---|---|
| | PSNR | SSIM | PSNR | SSIM | PSNR | SSIM | PSNR | SSIM |
| GAN-VGG-16x16 | 24.65 | 0.8650 | 22.43 | 0.8939 | 20.16 | 0.8617 | 25.70 | 0.9546 |
| GAN-ResNet-16x16 | 24.70 | 0.8672 | 22.48 | 0.8952 | 20.24 | 0.8622 | 25.92 | 0.9578 |
| GAN-VGG-ViT-16x16 | 24.90 | 0.8790 | 23.80 | 0.9150 | 20.96 | 0.8688 | 26.30 | 0.9640 |
| GAN-ResNet-ViT-16x16 | 25.80 | 0.9010 | 24.20 | 0.9220 | 21.62 | 0.8774 | 27.20 | 0.9714 |
| GAN-ResNet-ViT(1-Column) | 26.90 | 0.9115 | 25.50 | 0.9342 | 22.40 | 0.8990 | 28.30 | 0.9812 |
| **GAN-ResNet-ViT-(2-Column)** | **28.10** | **0.9270** | **26.90** | **0.9480** | **23.60** | **0.9090** | **29.98** | **0.9980** |
| GAN-ResNet-ViT(4-Column) | 26.10 | 0.9060 | 24.80 | 0.9290 | 22.10 | 0.8810 | 27.65 | 0.9752 |
| GAN-ResNet-ViT(1-Row) | 26.10 | 0.9030 | 24.65 | 0.9285 | 21.90 | 0.8795 | 27.45 | 0.9744 |
| GAN-ResNet-ViT(2-Row) | 27.98 | 0.9235 | 26.35 | 0.9435 | 23.25 | 0.9045 | 29.30 | 0.9910 |
| GAN-ResNet-ViT(4-Row) | 26.00 | 0.9610 | 24.90 | 0.9325 | 22.20 | 0.8880 | 27.35 | 0.9730 |

Moreover, we compare the results of the comparative models in two cases: with and without our pre-processing model. More specifically, Figs. 13-19 show the visual results of four comparative models on four public datasets (Pairs Street View, ImageNet, Places2, and CelebA-HQ). As these figures show, all of the comparative models have a better visual performance in the case of using proposed pre-processing methodology. Additionally, the numerical results are also reported in Table 3 using two evaluation metrics. As the results of this table show, all models have a better performance in the case of using the proposed pre-processing methodology due to capability of the proposed methodology in providing more informative features for the initial mask.

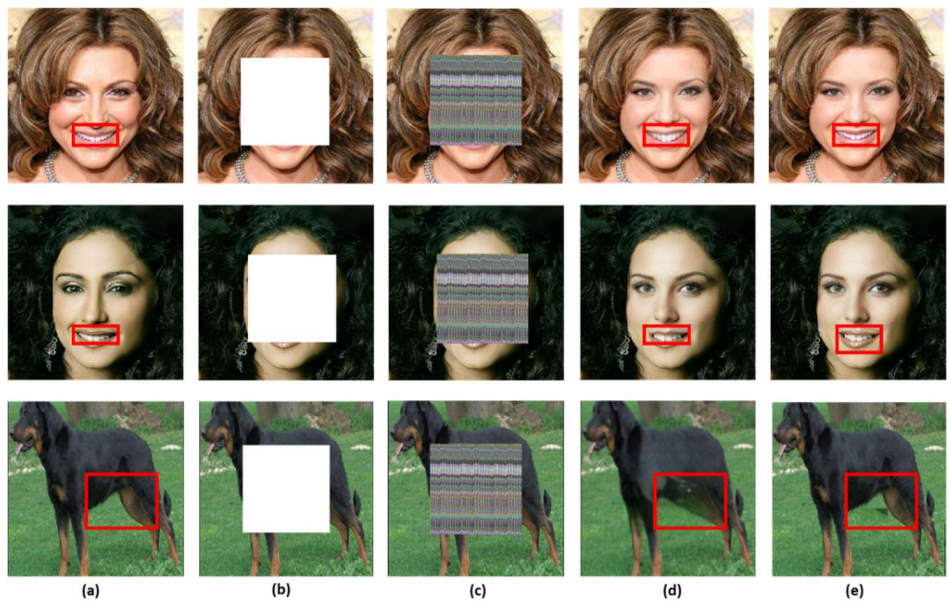

Fig. 13: Visual results of the proposed model: (a) Input image, (b) Masked image, (c) Masked image filled out with the ViT, (d) Reconstructed image using the base model, (e) Reconstructed image using the proposed methodology added to the GMCNN. The first, second, and third rows are corresponding to CelebA-HQ, CelebA-HQ, and ImageNet, respectively. Red boxes show the locations with major changes during the inpainting process.

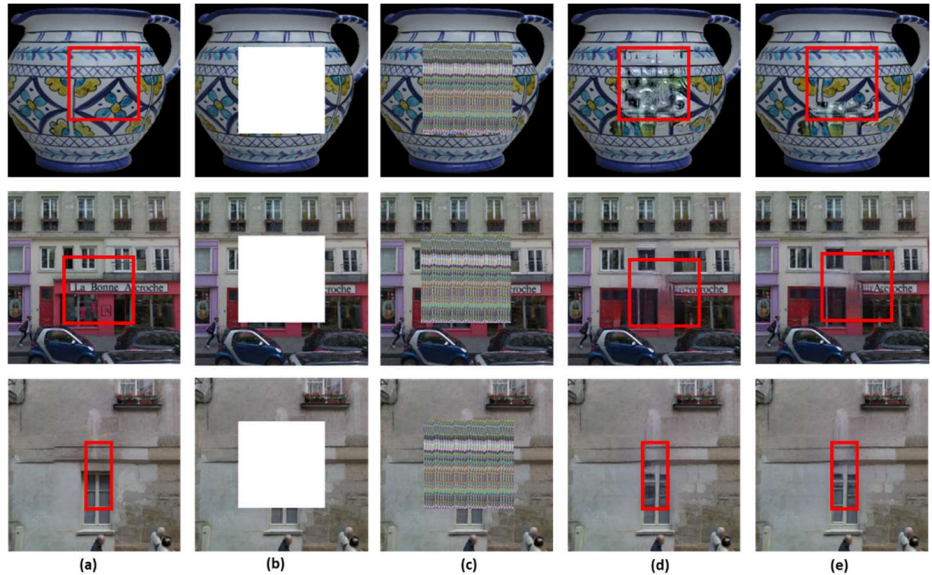

Fig. 14: Visual results of the proposed model: (a) Input image, (b) Masked image, (c) Masked image filled out with the ViT, (d) Reconstructed image using the base model, (e) Reconstructed image using the proposed methodology added to the GMCNN. The first, second, and third rows are corresponding to ImageNet, Paris Street View, and Paris Street View, respectively. Red boxes show the locations with major changes during the inpainting process.

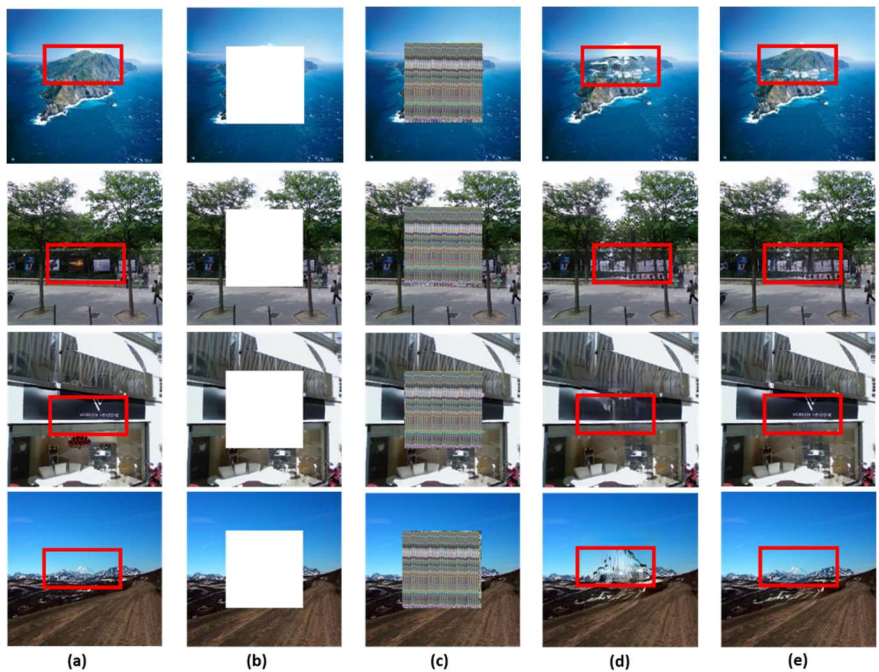

Fig. 15: Visual results of the proposed model: (a) Input image, (b) Masked image, (c) Masked image filled out with the ViT, (d) Reconstructed image using the base model, (e) Reconstructed image using the proposed methodology added to the GMCNN. The first, second, third, and fourth rows are corresponding to Places2, Paris Street View, Paris Street View, and Places2, respectively. Red boxes show the locations with major changes during the inpainting process.

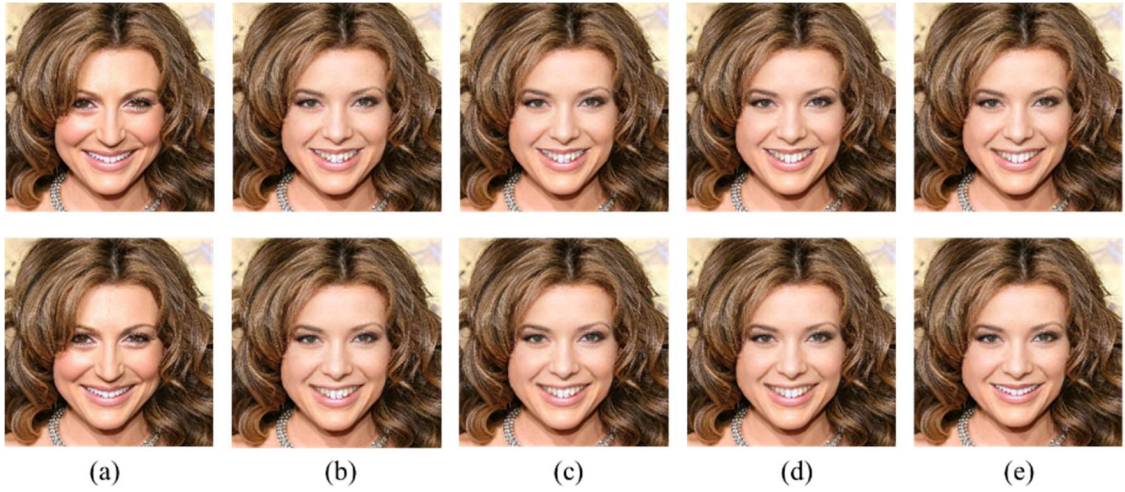

Fig. 16: Visual results of the proposed pre-processing methodology on the CelebA-HQ dataset: (a) Original image, (b) Obtained result from the CE, (c) result from the MNPAS, (d) Obtained result from the CA, and (e) Obtained result from the GMCNN. The first row is the results of the original models without considering our methodology and the Obtained second row is the results in the presence of our methodology.

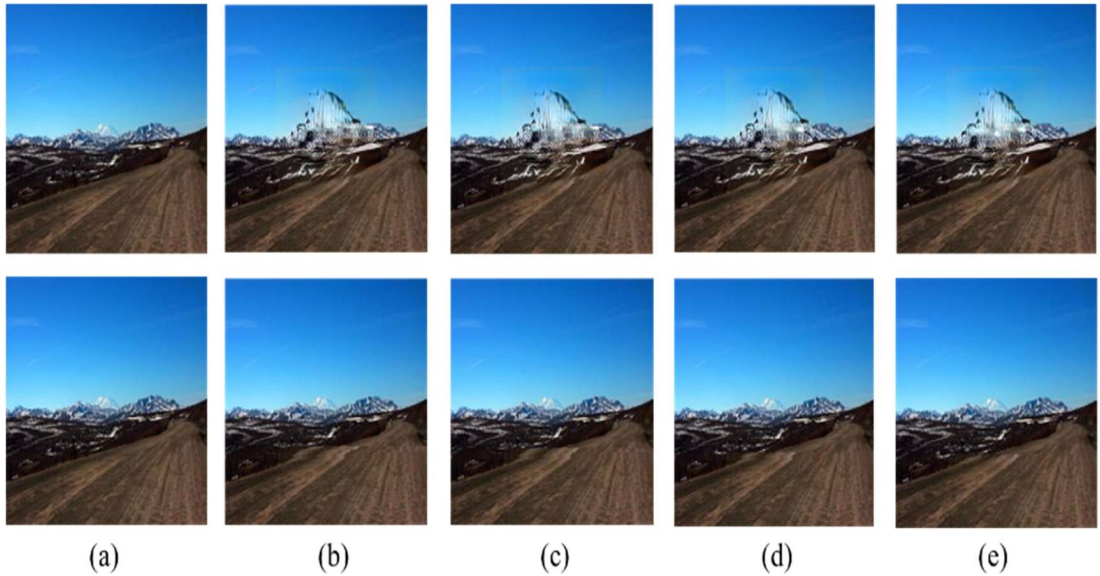

Fig. 17: Visual results of the proposed pre-processing methodology on the Place2 dataset: (a) Original image, (b) Obtained result from the CE, (c) Obtained result from the MNPAS, (d) Obtained result from the CA, and (e) Obtained result from the GMCN. The first row is the results of the original models without considering our methodology and the Obtained second row is the results in the presence of our methodology.

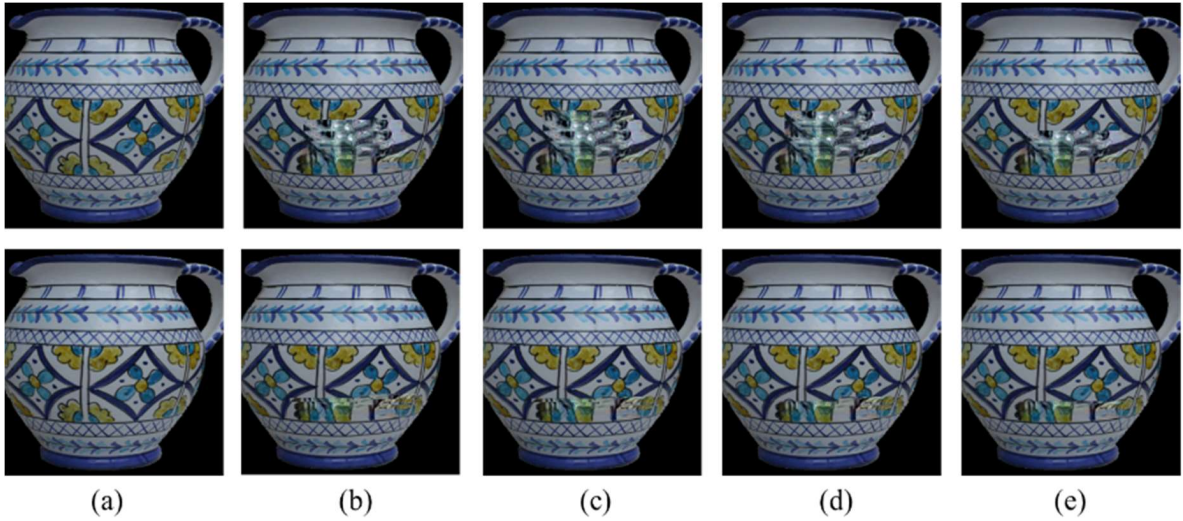

Fig. 18: Visual results of the proposed pre-processing methodology on the ImageNet dataset: (a) Original image, (b) Obtained result from the CE, (c) Obtained result from the MNPAS, (d) Obtained result from the CA, and (e) Obtained result from the GMCN. The first row is the results of the original models without considering our methodology and the Obtained second row is the results in the presence of our methodology.

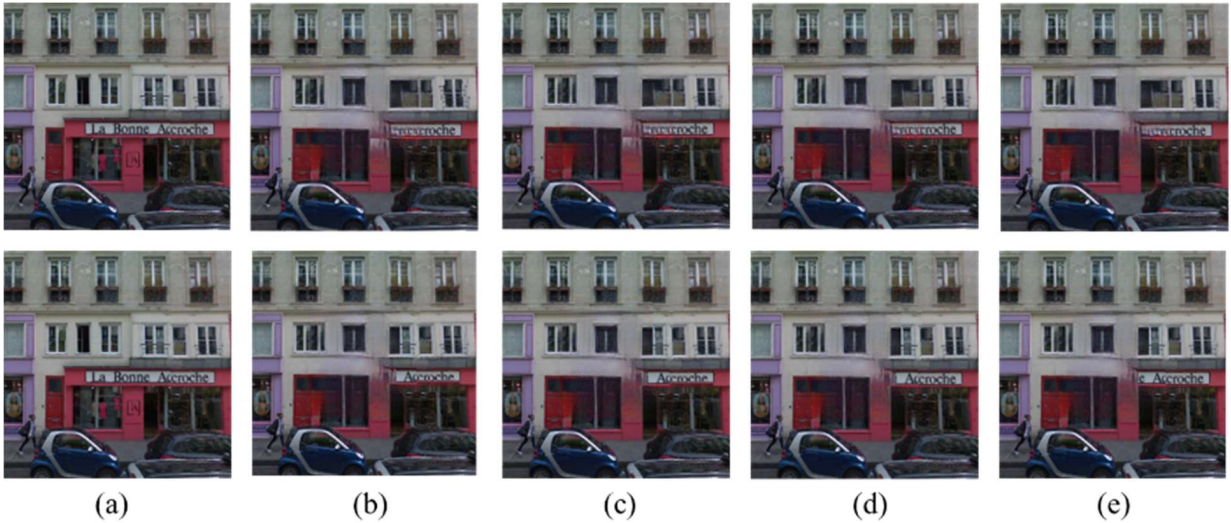

Fig. 19: Visual results of the proposed pre-processing methodology on the Pairs dataset: (a) Original image, (b) Obtained result from the CE, (c) Obtained result from the MNPAS, (d) Obtained result from the CA, and (e) Obtained result from the GMCN. The first row is the results of the original models without considering our methodology and the Obtained second row is the results in the presence of our methodology.

Table 3: Comparison the quantitative results of the proposed model with the SOTA models on four datasets for two distinct cases: with and without the proposed pre-processing methodology.

| Method | Pairs street view-100 | | ImageNet-200 | | Places2-2K | | CelebA-HQ-2K | |
|---|---|---|---|---|---|---|---|---|
| | PSNR | SSIM | PSNR | SSIM | PSNR | SSIM | PSNR | SSIM |
| CE (Deepak Pathak, 2016) | 22.10 | 0.8550 | 22.24 | 0.9010 | 17.20 | 0.8010 | 20.10 | 0.9002 |
| CE + Proposed pre-processing | 23.50 | 0.8734 | 23.58 | 0.9106 | 18.10 | 0.8115 | 21.30 | 0.9115 |
| MSNPS (Chao Yang, 2017) | 22.60 | 0.8560 | 22.30 | 0.9030 | 17.80 | 0.8080 | 20.60 | 0.9050 |
| MSNPS + Proposed pre-processing | 23.78 | 0.8788 | 23.64 | 0.9120 | 19.15 | 0.8220 | 21.90 | 0.9180 |
| CA (Jiahui Yu, 2018) | 22.90 | 0.8477 | 20.62 | 0.7217 | 18.20 | 0.8280 | 21.60 | 0.9260 |
| CA + Proposed pre-processing | 24.44 | 0.8590 | 22.65 | 0.9065 | 20.03 | 0.8439 | 23.90 | 0.9370 |
| GMCNN (Yi Wang, 2018) | 24.65 | 0.8650 | 22.43 | 0.8939 | 20.16 | 0.8617 | 25.70 | 0.9540 |
| GMCNN + Proposed pre-processing | 28.10 | 0.9270 | 26.90 | 0.9480 | 23.60 | 0.9090 | 29.80 | 0.9930 |

Finally, we discuss the impact of pre-training of the proposed pre-processing methodology. Since the original ViT uses the input patches with the dimension of 16x16, in the case of using the proposed pre-processing methodology, we need to train the ViT model with the pre-defined patch size. Once the model is trained in this way, the trained model can be used in the inference as a pre-trained model. In the case of using the original ViT model, we will be restricted to the input patch dimension of 16x16. Table 4 shows the results of the comparative models in three cases: without the proposed pre-processing, pre-training the ViT with the predefined patch size in the proposed pre-processing, and using the original ViT in the proposed pre-processing. As this table shows, compared to the original comparative models, using the proposed pre-processing leads to performance improvement in both cases of pre-training the ViT with the predefined patch size in the proposed pre-processing and using the original ViT in the proposed pre-processing. However, pre-training is led to the better performance.

Table 4: Numerical results of the comparative models in three cases: Original model, Original model plus the original ViT model with 16x16 patch size, and Original model plus the ViT model with a predefined patch size.

| Method | Pairs street view-100 | | ImageNet-200 | | Places2-2K | | CelebA-HQ-2K | |
|---|---|---|---|---|---|---|---|---|
| | PSNR | SSIM | PSNR | SSIM | PSNR | SSIM | PSNR | SSIM |
| CE (Deepak Pathak, 2016) | 22.10 | 0.8550 | 22.24 | 0.9010 | 17.20 | 0.8010 | 20.10 | 0.9002 |
| CE + ViT(16x16) | 22.80 | 0.8630 | 22.70 | 0.9040 | 17.50 | 0.8040 | 20.35 | 0.9025 |
| **CE + ViT(2-Column)** | **23.50** | **0.8734** | **23.58** | **0.9106** | **18.10** | **0.8115** | **21.30** | **0.9115** |
| MSNPS (Chao Yang, 2017) | 22.60 | 0.8560 | 22.30 | 0.9030 | 17.80 | 0.8080 | 20.60 | 0.9050 |
| MSNPS + ViT(16x16) | 22.96 | 0.8610 | 22.65 | 0.9060 | 18.15 | 0.8110 | 20.90 | 0.9090 |
| **MSNPS + ViT(2-Column)** | **23.78** | **0.8788** | **23.64** | **0.9120** | **19.15** | **0.8220** | **21.90** | **0.9180** |
| CA (Jiahui Yu, 2018) | 22.90 | 0.8590 | 22.65 | 0.9065 | 18.20 | 0.8280 | 21.60 | 0.9260 |
| CA + ViT(16x16) | 23.40 | 0 | | | | | | |
| **CA + ViT(2-Column)** | **24.44** | **0.8477** | **20.62** | **0.7217** | **20.03** | **0.8439** | **23.90** | **0.9370** |
| GMCNN (Yi Wang, 2018) | 24.65 | 0.8650 | 22.43 | 0.8939 | 20.16 | 0.8617 | 25.70 | 0.9540 |
| GMCNN + ViT(16x16) | 25.80 | 0.9010 | 24.20 | 0.9220 | 21.62 | 0.8774 | 27.20 | 0.9714 |
| **GMCNN + ViT(2-Column)** | **28.10** | **0.9270** | **26.90** | **0.9480** | **23.60** | **0.9090** | **29.80** | **0.9930** |

## 6. Conclusion and future work

In this paper, we proposed a new deep learning-based pre-processing methodology using the ViT model and and various visual patches in the image. In this way, ViT is used as a preprocessor for the input image to substitute the zero values in the missing areas with the attended values obtained from the ViT. To this end, different visual patches have been used in the input image, aiming to obtain the efficient spatial features. Relying on the self-attention mechanism in ViT, a feature map was obtained from the input image. Considering the task and the characteristics of the image data, different visual patches in the image can be used to obtain the features that we considered three kinds of visual patches in the input image (vertical, horizontal, and square) to construct the self-attention matrix. Experimental results using four comparative models on four public datasets confirm the efficacy of the proposed pre-processing methodology for image inpainting task. As a future work, we aim to employ the diffusion models for obtaining more efficient and robust features, leading to the better performance in image inpainting task.

**References**

Alain Horé, D. Z. (2010). Image Quality Metrics: PSNR vs. SSIM. *20th International Conference on Pattern Recognition Date of Conference, Istanbul, Turkey*.

Alexei A. Efros, T. K. (1999). Texture synthesis by nonparametric sampling. *Proceedings of the Seventh IEEE International Conference on Computer Vision* (pp. 1033–1038). IEEE.

Alexei A. Efros, W. T. (2001). Image quilting for texture synthesis and transfer. *Proceedings of the 28th annual conference on Computer graphics and interactive techniques* (pp. 341–346). ACM.

C. Ballester, M. B. (2001). Filling-in by joint interpolation of vector fields and gray levels. *IEEE transactions on image processing 10*, 1200–1211.

Chao Yang, X. L. (2017). High-resolution image inpainting using multi-scale neural patch synthesis. *IEEE Conference on Computer Vision and Pattern Recognition (CVPR)* (p. 3). Honolulu, HI, USA: IEEE.

Cho, T. S., Butman, M., Avidan, S., & Freeman, W. T. (2008). The patch transform and its applications to image editing. *IEEE Conference on Computer Vision and Pattern Recognition.* Anchorage, AK, USA.

Chuan Li, M. W. (2016). Combining markov random fields and convolutional neural networks for image synthesis. *Proceedings of the IEEE Conference on Computer Vision and Pattern Recognition (CVPR)*, 2479–2486.

Connelly Barnes, E. S. (2009). Patchmatch: A randomized correspondence algorithm for structural image editing. *ACM Transactions on Graphics (TOG) 28*, 1-11.

Dan Ciregan, U. M. (2012). Multi-column deep neural networks for image classification. *IEEE Conference on Computer Vision and Pattern Recognition (CVPR)*, 3642–3649.

Deepak Pathak, P. K. (2016). Context encoders: Feature learning by inpainting. *Proceedings of the IEEE Conference on Computer Vision and Pattern Recognition*, (pp. 2536–2544). Las Vegas, NV, USA.

Denis Simakov, Y. C. (2008). Summarizing visual data using bidirectional similarity. *IEEE Conference on Computer Vision and Pattern Recognition (CVPR), Anchorage, AK, USA*.

Forest Agostinelli, M. R. (2013). Adaptive multi-column deep neural networks with application to robust image denoising. *Advances in Neural Information Processing Systems (NIPS) 26*, 1493–1501.

Haiwei Wu, J. Z. (2021). Deep Generative Model for Image Inpainting with Local Binary Pattern Learning and Spatial Attention. *IEEE Transactions on Multimedia 24*, 4016-4027.

ImgNet (2024). Retrieved from: https://www.image-net.org/

Ishaan Gulrajani, F. A. (2017). Improved training of wasserstein gans. *NIPS*, (pp. 5769–5779). Monreal, Canada.

Jiahui Yu, Z. L. (2018). Generative image inpainting with contextual attention. *Proceedings of the IEEE Conference on Computer Vision and Pattern Recognition (CVPR)*, 5505-5514.

Jifeng Dai, H. Q. (2017). Deformable convolutional networks. *IEEE International Conference on Computer Vision (ICCV), Venice, Italy*.

Kaiming He, J. S. (2012). Statistics of patch offsets for image completion. *European Conference on Computer Vision (ECCV), Florence, Italy*, 16-29.

Kaiming He, J. S. (2014). Image completion approaches using the statistics of similar patches. *IEEE Transactions on Pattern Analysis and Machine Intelligence 36*, 2423–2435.

Kaiming He, X. Z. (2016). Deep residual learning for image recognition. *Proceedings of the IEEE conference on computer vision and pattern recognition* (pp. 770–778). Las Vegas, NV, USA: IEEE.

Li Xu, J. S. (2014). Deep convolutional neural network for image deconvolution. *Advances in Neural Information Processing Systems (NIPS) 27*, 1790–1798.

Marcelo Bertalmio, G. S. (2000 ). Image inpainting. *SIGGRAPH '00: Proceedings of the 27th annual conference on Computer graphics and interactive techniques*, 417 – 424.

Omar Elharrouss, R. D. (n.d.). Transformer-based Image and Video Inpainting: Current Challenges and Future Directions. *arXiv:2407.00226*, 2024.

Places2 (2024). Retrieved from: http://places2.csail.mit.edu/download-private.html

PyTorch. (2024, Feb.). Retrieved from PyTorch: https://pytorch.org

Raymond A. Yeh, C. C.-J. (2017). Semantic image inpainting with perceptual and contextual losses. *Proceedings of the IEEE Conference on Computer Vision and Pattern Recognition.* Honolul, Hawaii, USA: IEEE.

Rolf Köhler, C. S. (2014). Mask-specific inpainting with deep neural network. *German Conference on Pattern Recognition*, 523–534.

Satoshi Iizuka, E. S.-S. (2017). Globally and locally consistent image completion. *ACM Transactions on Graphics (TOG) 36*, 1-14.

Snelgrove, X. (2017). High-resolution multi-scale neural texture synthesis. *SIGGRAPH ASIA 2017 Technical Briefs. ACM*, 1-4.

T. Karras, T. A. (2018). Progressive growing of gans for improved quality, stability, and variation. *The Sixth International Conference on Learning Representations (ICLR)*.

Weize Quan, J. C.-M. (2024). Deep Learning-based Image and Video Inpainting: A Survey. *International Journal of Computer Vision 132*, 2367–2400.

Wenbo Li, Z. L. (2022). MAT: Mask-Aware Transformer for Large Hole Image Inpainting. *IEEE/CVF Conference on Computer Vision and Pattern Recognition (CVPR), New Orleans, LA, USA*, 10758-10768.

Xiaobo Zhang, D. Z. (2023). Image inpainting based on deep learning: A review. *Information Fusion 90*, 74-94.

Yi Wang, X. T. (2018). Image Inpainting via Generative Multi-column Convolutional Neural Networks. Montreal, Canada.

Yijun Li, S. L.-H. (2017). Generative face completion. *IEEE Conference on Computer Vision and Pattern Recognition (CVPR)* (pp. 5892-5900). Honolulu, Hawaii, USA: IEEE.

Yingying Zhang, D. Z. (2016). Single-image crowd counting via multi-column convolutional neural network. *IEEE Conference on Computer Vision and Pattern Recognition (CVPR)*, 589–597.

Yunho Jeon, J. K. (2017). Active convolution: Learning the shape of convolution for image classification. *IEEE Conference on Computer Vision and Pattern Recognition (CVPR), Honolulu, HI, USA*.